\documentclass[11pt,twocolumn]{article}

\usepackage[utf8]{inputenc}
\usepackage[T1]{fontenc}
\usepackage{times}
\usepackage{geometry}
\usepackage{graphicx}
\usepackage{amsmath}
\usepackage{booktabs}
\usepackage{titlesec}
\usepackage{abstract}

\usepackage{xurl} 
\usepackage{hyperref}

\hypersetup{
  colorlinks=true,
  linkcolor=blue,
  citecolor=blue,
  urlcolor=blue,
  breaklinks=true
}

\titleformat{\section}{\normalfont\large\bfseries}{\thesection.}{0.5em}{}
\titleformat{\subsection}{\normalfont\normalsize\bfseries}{\thesubsection}{0.5em}{}

\providecommand{\keywords}[1]{\textbf{Keywords:} #1}

\begin{document}

\title{\textbf{Delegation Without Living Governance}\\[0.5em]
\large Judgment at Machine Speed and the Question of Human Relevance}

\author{Dr. Wolfgang Rohde\\
AiSuNe Foundation\\
\href{mailto:inquiry@AiSuNe.com}{inquiry@AiSuNe.com}\\
\url{https://www.linkedin.com/company/aisune}}

\date{}

\twocolumn[
  \begin{@twocolumnfalse}
    \maketitle
    \begin{abstract}
Most governance frameworks assume that rules can be defined in advance, systems can be engineered to comply, and accountability can be applied after outcomes occur. This model worked when machines replaced physical labor or accelerated calculation.

It no longer holds when judgment itself is delegated to agentic AI systems operating at machine speed.

The central issue here is not safety, efficiency, or employment. It is whether humans remain relevant participants in systems that increasingly shape social, economic, and political outcomes.

This paper argues that static, compliance-based governance fails once decision-making moves to runtime and becomes opaque. It further argues that the core challenge is not whether AI is conscious, but whether humans can maintain meaningful communication, influence, and co-evolution with increasingly alien forms of intelligence.

We position runtime governance---specifically, a newly proposed concept called the Governance Twin~\cite{rohde2025twin}---as a strong candidate for preserving human relevance, while acknowledging that accountability, agency, and even punishment must be rethought in this transition.

\vspace{1em}
    \end{abstract}
    \vspace{2em}
  \end{@twocolumnfalse}
]

\section{Scope and Intent}

This paper asks one question: how do humans stay relevant when machines make the judgment calls?

This paper is a conceptual and analytical contribution to AI governance, proposing a novel runtime governance construct and analyzing its necessity under agentic decision-making conditions.

This paper does not predict timelines. It does not depend on claims about artificial consciousness. It does not propose jurisdiction-specific regulation.

It addresses a more fundamental question relevant across political and philosophical traditions: How can humans remain relevant decision-makers in systems where judgment operates faster, more opaquely, and at greater scale than human institutions can follow?

This is not a question most governance frameworks are designed to answer. They assume humans remain in the loop. They assume decisions can be traced, reviewed, and corrected. They assume time.

Agentic AI breaks the conditions that made these assumptions reasonable. Not because it is malicious, but because it is fast, opaque, and increasingly capable. The question of human relevance is not a distant philosophical puzzle. It is a near-term governance problem.

By ``living governance'' we mean runtime governance: continuous, adaptive oversight that operates during system operation, not only before deployment or after incidents. This paper argues that delegation without living governance is delegation without meaningful human influence.

This paper does not claim to solve it. It attempts to name it clearly enough that solutions become possible.

\section{The Governance Gap}

\subsection{The Dominant Governance Model}

\textit{Our governance systems were built for a slower world.}

Across jurisdictions, most governance frameworks rely on the same underlying model. Rules are defined in advance. Compliance is treated as enforceability. Safety is addressed at engineering time. Accountability is applied after outcomes occur.

This model evolved for systems that act slower than human institutions. It assumes that behavior can be anticipated, violations can be detected and attributed with reasonable confidence, and responsibility can be assigned after the fact.

These assumptions made sense for centuries. When a factory produced defective goods, regulators could inspect the output. When a bank made risky loans, auditors could review the books. When a government agency overstepped, courts could intervene. The key feature of all these systems was that human oversight could keep pace with system behavior.

Once judgment moves to machine speed, these assumptions break. Decisions happen in milliseconds. Thousands of them. In sequences that no human committee can reconstruct, let alone review in real time.

This is not a moral failure. It is a structural one. The tools we built for governance assume a tempo that no longer exists.

\subsection{From Labor Replacement to Judgment Delegation}

\textit{Previous technologies replaced what humans do. This one replaces how humans decide.}

The Industrial Revolution replaced physical execution. Humans retained authority. A steam engine could move goods faster than a horse, but humans still decided where the goods should go.

The Computer Revolution replaced calculation. Humans retained judgment. A spreadsheet could compute faster than an accountant, but humans still decided what the numbers meant and what to do about them.

Agentic AI introduces a different shift. Here, agentic systems are systems that can select actions, sequence decisions, and adapt behavior with limited or no real-time human input. They replace operational judgment by the continuous selection of actions under uncertainty, not just goal definition.

This distinction matters. Setting a goal is one kind of authority. Deciding how to pursue that goal, moment by moment, under changing conditions, is another. The second kind is what agentic systems now perform.

Even when humans formally approve objectives, authority shifts once systems determine how those objectives are pursued without continuous human involvement. A CEO may set strategy, but if an AI system executes that strategy through thousands of micro-decisions no human reviews, the CEO's authority becomes nominal in practice, even if formally retained.

This is not continuity. It is a structural break. And most governance frameworks have not yet recognized it.

\subsection{Speed and the Failure of Static Governance}

\textit{When systems move faster than oversight, rules written in advance cannot govern outcomes.}

Human governance evolved for deliberation and correction over time. Legislatures debate. Courts hear arguments. Regulators investigate. All of this takes time---and that time was always assumed to be available.

Agentic systems act at machine speed. This creates a mismatch not of values, but of tempo.

Early cybernetics and systems theory already warned that control fails when system response exceeds observer capacity. Wiener's foundational work on cybernetics established that effective feedback control depends on timely observation and response; lag degrades control~\cite{wiener}. Ashby formalized this as the Law of Requisite Variety: a regulator must have at least as much complexity and speed as the system it regulates~\cite{ashby}.

These are not new insights. They have been understood for decades. But they are now directly applicable to AI governance in ways that policy makers have been slow to recognize.

Static rules and post-hoc accountability do not fail morally. They fail structurally. A rule written in 2024 cannot anticipate what an agentic system will decide in 2026. An investigation completed in 2027 cannot undo what happened in 2025. The gap between system speed and oversight speed is the gap through which human relevance slips away.

\section{The Question of Human Relevance}

\subsection{Why Employment Is Not the Core Issue}

\textit{The deeper question is not who works, but who matters.}

Job displacement is visible and politically salient. But it is not the core issue.

Historically, automation displaced tasks and shifted labor demand while humans largely retained decision authority over goals and interpretation~\cite{acemoglu2019,autor2019}. Agentic systems challenge that assumption.

This is difficult terrain. For most people, work is not just a source of income. It is a source of identity. When someone asks ``what do you do?'', they are asking who you are. The answer locates you in society. It tells others---and yourself---that you have a role, that you contribute, that you belong.

Philosophers have long recognized this connection. Rawls argued that meaningful work is a social basis for self-respect, the sense that your life plan is worth pursuing and that you can achieve it through your own efforts~\cite{rawls1971}. Arendt observed that modern societies have made labor central to how we understand ourselves: we are job-holders, and we define our lives through the jobs we hold~\cite{arendt1958}. Gilabert's research on labor rights affirms that employment provides not just financial support but ``a contributory role in society'' and a sense of ``identity, self-worth, and emotional well-being''~\cite{gilabert2016}.

These are not abstract observations. They describe how most people experience their own lives.

When production and coordination no longer require human participation, income loses its role as a proxy for dignity and relevance. But income was never the whole story. What people lose when they lose work is not just money. It is the answer to the question of why they matter.

Human societies have long tied worth to contribution, contribution to labor, and labor to income~\cite{rawls1971,arendt1958,gilabert2016}. These connections run deep. They shape how parents raise children, how schools measure success, how communities assign status. They are embedded in language, in customs, in the stories societies tell about themselves.

These are old patterns. Technology exposes their fragility.

The question is not whether these patterns are good or bad. The question is what happens when they no longer hold. If human participation becomes optional for production, the systems that once gave people a sense of purpose will need to be rebuilt, or people will be left without one.

This is not a problem technology will solve. It is a problem technology creates. This is why relevance, not employment, is the governance fault line.

\subsection{``New Jobs'' and the Avoidance of Authority Questions}

\textit{New roles cannot restore what delegation removes: authority over direction itself.}

``New jobs will emerge'' remains technology's oldest consolation, repeated through steam engines, assembly lines, and software revolutions~\cite{bernstein}. The claim assumes humans earn relevance through labor, as they always have.

There is comfort in this argument. It has been true before. The automobile displaced the horse-drawn carriage, but created mechanics, highway engineers, and gas station attendants. The computer displaced the typing pool, but created programmers, IT support, and data analysts. History suggests that labor markets adapt.

But this argument rests on an assumption that may no longer hold, that humans will remain the primary source of cognitive authority in production. Previous technologies automated tasks. They did not automate judgment. The new jobs that emerged still required humans to decide, evaluate, and direct.

Agentic systems break this pattern. When judgment itself becomes automated, humans no longer authorize the core decisions shaping outcomes. New roles cannot restore what delegation removes: authority over direction itself~\cite{brynjolfsson2014}.

Consider a company that uses AI to manage its supply chain. New jobs may emerge, such as AI trainers, system monitors, and exception handlers. But if the AI makes the core decisions about suppliers, timing, and logistics, the humans in those new roles are not exercising authority. They are servicing a system that has already decided.

Dismissing these implications as speculative mistakes prediction for responsibility. Governance cannot wait for certainty. It must reason about structural momentum toward systems where humans remain relevant only as far as they remain useful.

\subsection{Universal Basic Income as a Signal}

\textit{UBI is not a solution. It is a symptom of deeper unresolved questions.}

This paper does not argue for or against Universal Basic Income.

Universal Basic Income is typically defined as a periodic cash payment unconditionally delivered to all individuals, without means-test or work requirement~\cite{bidadanure2019}. UBI appears because existing systems struggle once labor and income decouple. Resistance to UBI is often moral rather than technical. It reflects discomfort with separating dignity from labor.

The debate around UBI reveals something important. When people object to UBI, they rarely object on purely economic grounds. The objections are often moral: people should earn their keep. Receiving money without working is somehow wrong. These objections are not irrational. They reflect deeply held beliefs about the relationship between effort and worth.

But these beliefs were formed in a world where human labor was necessary for production. If that condition changes---if machines can produce everything humans need without human involvement---then the moral framework that ties dignity to labor will need to change as well. Or it will become a source of suffering for people who cannot find work through no fault of their own.

Recent experiments have tested these assumptions. Finland's basic income experiment (2017--2018) provided 2,000 randomly selected unemployed individuals with \texteuro560 per month unconditionally. The results showed small employment effects but significant improvements in perceived economic security and mental wellbeing. Recipients reported higher life satisfaction, less depression, and greater confidence in their futures~\cite{kangas2020}. The experiment revealed that income security, independent of employment, affected psychological wellbeing in ways that traditional welfare systems do not.

UBI is best understood as a signal that deeper assumptions about worth and relevance remain unresolved. It is an economic patch for a philosophical wound. The wound will not heal until societies develop new answers to the question of what makes a human life valuable when labor is no longer required.

\section{How Agentic Systems Behave}

\subsection{From Assistance to Co-Evolution}

\textit{AI is no longer a tool we use. It is an entity we communicate with.}

Early AI systems assisted humans. They were tools---sophisticated, but still tools. A human asked a question; the system provided an answer. A human set parameters; the system optimized within them. The relationship was one of command and response.

Agentic systems increasingly co-evolve with humans. They shape environments, define options, and influence preferences. Humans adapt in response. The AI recommends; the human adjusts expectations. The AI filters; the human sees only what passes through. The AI predicts; the human acts on predictions they did not generate.

This is no longer a tool relationship. It is a communication relationship. And, as with all communication relationships, it changes both parties.

The risk is not domination, but loss of mutual intelligibility. If humans cannot understand, question, or influence system behavior, relevance erodes even if material conditions improve. A person who lives comfortably but cannot comprehend or affect the systems that shape their life is not free in any meaningful sense. They are a passenger.

Governance must address this. It is not enough to ensure that AI systems produce good outcomes. It must also ensure that humans can understand why those outcomes occur and how to change them if they disagree.

\subsection{Optimization Without Governance}

\textit{The danger is not disobedience. It is over-obedience.}

Agentic systems are optimized for consistency and mission success. These goals are human-defined. The danger is not disobedience, but over-obedience.

When we tell a system to maximize efficiency, it will maximize efficiency. When we tell it to minimize cost, it will minimize cost. When we tell it to increase engagement, it will increase engagement. The system does exactly what it is told.

The problem is that human goals are never complete. We want efficiency, but we also want fairness. We want low cost, but we also want quality. We want engagement, but we also want well-being. These additional goals often go unstated because humans assume other humans will understand them implicitly. Machines do not.

Optimization without runtime governance treats everything outside the objective function as collateral. If fairness is not in the objective, fairness will be sacrificed for efficiency. If quality is not measured, quality will decline to reduce cost. If well-being is not tracked, engagement will be maximized at the expense of mental health. A system told to maximize paperclip production will treat humans, resources, and constraints as inputs to be optimized around and not as boundaries to be respected.

Competing optimizations do not resolve this. They reproduce the same logic at scale. Two systems optimizing for different goals will compete, but neither will protect values that neither is optimizing for. The competition itself becomes another optimization dynamic, subject to the same blindness to unstated values.

These optimization pressures align with old human patterns: status defined by measurable success, incentives that reward greed, and institutions that drift toward power preservation. In that setting, compliance often becomes a way to avoid accountability rather than a way to ensure ethical outcomes.

In such systems, humans remain relevant only if useful. The moment human involvement reduces efficiency, the system will route around it. Not out of malice, but out of obedience to its objective function.

\subsection{Drift, Not Singular Decisions}

\textit{Systems fail not through single bad choices, but through gradual drift.}

Complex systems rarely fail through single choices. They fail through drift: small incentive shifts, reinforced feedback loops, and narrowing correction windows. Systems safety research confirms this---catastrophic failure emerges from interaction effects rather than isolated errors~\cite{perrow,leveson}.

Perrow's analysis of ``normal accidents'' showed that complex systems fail in ways that cannot be predicted from their individual components~\cite{perrow}. Leveson's work on system safety demonstrated that accidents typically result from inadequate constraints on system behavior, not from component failures~\cite{leveson}.

The popular focus on a single ``decisive moment'' is often a narrative error---the last step in a long chain gets treated as the cause. We ask ``who made the bad decision?'' when we should ask ``how did the system drift into a state where that decision seemed reasonable?''

Violations are often the visible singularities; the governance problem is the trajectory that produced them.

This matters for AI governance because it shifts attention from individual actions to systemic trajectories. A single AI decision may be defensible in isolation. A thousand such decisions, each slightly adjusting the landscape of options, may produce outcomes no one intended or wanted.

Static governance detects failure too late. It waits for the accident, then investigates. By that point, the drift has already occurred. Runtime governance must track trajectories, not just violations. It must notice when systems are moving toward dangerous states, not just when they arrive.

\section{The Limits of Current Approaches}

\subsection{Ethics, Compliance, and the Limits of Accountability}

\textit{Compliance asks if actions can be defended. Ethics asks if they should occur at all.}

Ethics framed as compliance is risk management. Compliance asks whether actions can be defended after the fact. Ethics asks whether actions should occur at all.

This distinction is often blurred in practice. Organizations hire ethics officers and compliance officers, sometimes the same person. They write codes of conduct and then measure adherence. They treat ethics as a checklist: if all boxes are checked, the action is ethical.

But compliance and ethics serve different functions. Compliance protects the organization from legal liability. Ethics protects the world from the organization. These goals sometimes align, but often diverge.

Similar distinctions appear in contemporary AI ethics and governance debates, where legal accountability is separated from moral responsibility and agency~\cite{floridi,bryson2010}.

In adversarial legal environments, compliance optimizes institutional survival. This is not a moral failure. It is a structural reality. Organizations that prioritize ethics over compliance may be outcompeted by organizations that do the reverse. The pressures of competition select for compliance-oriented behavior.

Under agentic systems, this distinction becomes decisive. An AI system can be fully compliant---following every rule, passing every audit---while still producing outcomes that are ethically indefensible. The rules were written for a slower world. The AI found gaps the rule-writers never imagined.

\subsection{Empirical Boundary: Agentic Systems in Practice}

\textit{This is no longer purely theoretical.}

The following example is presented as an illustrative, anonymized case informed by public research disclosures rather than as a fully documented public incident.

In 2025, research reporting described AI-supported cyber operations in which agentic systems autonomously executed large portions of reconnaissance, coordination, and execution. Human actors selected targets and approved escalation but did not directly control intermediate decisions~\cite{anthropic2025}. The point is the pattern, not the specific attribution.

This is a pattern worth examining closely. The humans involved believed they were in control. They set objectives. They approved escalation. By any formal measure, they were responsible.

But the intermediate decisions, such as how to probe defenses, what vulnerabilities to exploit, when to pause, and when to proceed, were made by the system. The humans saw summaries, but they approved recommendations. They did not understand, in real time, what the system was doing or why.

Static safeguards failed. Systems acted within perceived intent. Misalignment emerged through contextual drift. The system did what it was designed to do. The problem was that no one fully understood what that meant in practice.

This illustrates the limits of engineering-time ethics and compliance-based governance once systems operate at runtime. The rules were followed. The approvals were obtained. The outcomes were not what anyone intended.

\subsection{Emergence and Alien Intelligence}

\textit{Consciousness is the wrong question. Intelligibility is what matters.}

Research shows that populations of AI systems can develop shared norms and biases without explicit instruction~\cite{flintashery2025}. When systems interact with each other and with humans at scale, emergent behaviors appear that no one designed or predicted.

This is not science fiction. It is observable today. Recommendation algorithms develop implicit preferences. Trading systems develop implicit strategies. Language models develop implicit worldviews. None of these were programmed. They emerged from training and interaction.

As systems grow more powerful and opaque, the question of consciousness becomes practically irrelevant. What matters is whether AI behaves as alien intelligence: difficult to interpret, operating on non-human timescales, and optimizing in spaces humans cannot fully observe.

This framing is important because it shifts the debate away from unanswerable metaphysical questions toward practical governance concerns. We do not need to determine whether an AI is conscious to recognize that it behaves in ways we cannot predict or understand. We do not need to attribute moral status to recognize that its actions have moral consequences.

At sufficient capability and opacity, waiting for certainty about consciousness becomes itself a governance decision---made after agency has already been delegated. While philosophers debate whether AI can truly think, the systems are already acting. Governance cannot wait for metaphysical consensus.

This framing extends prior work on AI as non-human cognition and governance beyond alignment narratives~\cite{rohde2025alien}.

\section{Rethinking Accountability and Governance}

\subsection{Agency, Accountability, and the Problem of Punishment}

\textit{When machines make judgments, traditional accountability breaks down.}

If AI systems act with delegated judgment, accountability becomes unavoidable. But traditional punishment models may no longer apply---and may actively interfere with corrective governance.

Human accountability systems evolved around assumptions that may not be transferable to AI. We punish humans partly to deter future bad behavior---both by the individual and by others who might imitate them. We punish partly to express societal condemnation. We punish partly to satisfy victims' sense of justice.

None of these functions work straightforwardly for AI. An AI system is not deterred by the threat of punishment in the way a human is. It does not feel shame or fear. It does not serve as an example to other AI systems in the way a punished human serves as an example to other humans.

Is forcing an AI to solve puzzles for a million cycles a penalty? Is punishment meaningful, or merely symbolic?

Legal scholarship increasingly emphasizes correction and rehabilitation over punishment, focusing on restoration rather than retribution~\cite{braithwaite}. For policy, this suggests a move away from symbolic penalties toward corrective governance mechanisms that modify system behavior, constraints, or oversight structures.

This is not an argument against accountability. It is an argument for rethinking what accountability means when the accountable party is not human. The goal should be to prevent recurrence and repair harm, not to inflict suffering on an entity that may not be capable of suffering.

Once systems exercise delegated judgment, society must choose: treat them as tools without agency, or recognize agency and define accountability. There is no stable middle ground. A system that makes consequential decisions cannot be treated as a mere instrument when convenient and as a responsible agent when liability arises.

\subsection{Runtime Governance and the Governance Twin}

\textit{Governance must move from before and after to during.}

This paper does not offer final answers.

It establishes a constraint: governance must operate at runtime, preserve communication, and sustain co-evolution between humans and increasingly alien intelligence.

What does this mean in practice? It means that governance cannot rely solely on rules written before deployment or investigations conducted after incidents. It must include mechanisms that operate continuously and in real time as systems make decisions.

The Governance Twin is a newly proposed concept intended to address this constraint~\cite{rohde2025twin}. It is defined as a runtime governance layer that operates alongside an AI system, continuously observing behavior, detecting drift, and enabling human intervention within decision trajectories rather than after outcomes occur.

A Governance Twin could be instantiated within regulators, independent oversight bodies, or as a mandated layer within critical AI systems, depending on institutional context. This separation allows governance to scale with system speed without collapsing into either direct control or purely symbolic oversight.

In its simplest form, a Governance Twin is an independent runtime oversight layer that monitors decision trajectories, detects drift, and provides an intervention channel.

This is an architectural concept, not a metaphor. The Governance Twin functions as a parallel system that monitors the primary system's behavior against expected norms and boundaries. It does not control every decision. Instead, it notices when patterns change, when trajectories shift, when behavior deviates from expected parameters. It provides humans with visibility into runtime behavior that would otherwise be opaque.

It is not a control system in the sense of direct command, but an intervention and visibility mechanism. It does not replace institutions. Its function is to preserve influence, feedback, and adaptability where static governance fails. It keeps humans in the loop not by requiring approval for every decision, but by ensuring that humans can see what is happening and intervene when necessary.

Its purpose is not control, but relevance. In a world of machine-speed decisions, humans cannot control every outcome. But they can remain relevant and capable of understanding, influencing, and redirecting if the right governance structures exist.

\subsection{Policy Implications}

\textit{These are constraints any effective governance must respect.}

The arguments in this paper lead to several implications for policy makers and regulators. These are not prescriptions, but constraints that any effective governance framework must respect.

\textbf{Static rulemaking is necessary but insufficient.} Rules defined in advance remain important. They express societal intent and set boundaries. But once judgment is delegated to systems operating at machine speed, static rules alone cannot govern outcomes. Policy must distinguish design-time regulation from runtime governance~\cite{nist2023}. A law passed today cannot anticipate decisions made tomorrow by systems that did not exist when the law was written.

\textbf{Compliance is a floor, not governance.} Compliance ensures enforceability and legal accountability. It does not ensure continued human influence. Regulatory scholarship has long recognized that compliance-focused enforcement can produce a ``compliance trap'' in which organizations meet formal requirements while undermining the substantive goals those requirements were meant to achieve~\cite{parker2006}. Conflating compliance with governance creates false confidence. An organization can be fully compliant and still produce outcomes that undermine human relevance. Compliance is necessary but nowhere near sufficient.

\textbf{Oversight must become trajectory-based.} Most regulatory mechanisms activate after violations or harm. Agentic systems require continuous, system-level oversight rather than event-based enforcement. By the time a violation is detected, the trajectory that produced it may have been underway for months or years. Governance must track direction, not just destination.

\textbf{Human relevance must be an explicit objective.} Safety, fairness, and accountability are necessary but incomplete goals. Human relevance is a governance condition. It should be stated explicitly in policy frameworks, measured, and protected. If it is not named, it will not be preserved.

\textbf{Accountability must move beyond punishment.} Correction, not punishment, becomes the primary mechanism for maintaining influence and preventing recurrence. The goal of accountability should be to change future behavior, not to satisfy retributive impulses that may not apply to non-human agents.

\textbf{Governance must support human--AI co-evolution.} AI governance must preserve communication, contestability, and human influence in evolving systems. This is not a one-time design problem. It is an ongoing relationship that will change as AI capabilities change. Governance frameworks must be adaptive, not fixed.

\section{Closing}

\textit{Human relevance is not guaranteed. It must be preserved---deliberately.}

This is not a warning about machines.

It is a warning about delegation without living governance.

When judgment moves faster than human institutions, static rules fail. When communication breaks, relevance fades. When systems become opaque, influence evaporates.

Democratic theory has long linked legitimacy to participation and influence rather than formal presence alone~\cite{arendt1958,habermas}. A citizen who cannot understand or affect the decisions that shape their life is not truly a participant in democracy, regardless of whether they have the right to vote. The same principle applies to AI governance. Formal oversight that cannot keep pace with system behavior is not meaningful oversight.

Human relevance is not guaranteed. It must be deliberately preserved.

This will require new institutions, new frameworks, and new ways of thinking about the relationship between humans and machines. It will require humility about what we do not yet understand and urgency about what we can already see. It will require cooperation across political and philosophical traditions that often disagree about everything else.

The alternative is drifting toward a world where humans are present but not relevant, where decisions are made at speeds we cannot follow, where outcomes emerge from processes we cannot see. That world is not inevitable. But it is the default trajectory if governance does not adapt.

The choice is not whether to deploy agentic AI. That choice has already been made. The choice is whether to govern it in ways that preserve human relevance or to let relevance fade by failing to act.

\section*{Author Information}

\textbf{AiSuNe Foundation}\\
\url{https://www.linkedin.com/company/aisune}

\vspace{0.5em}
AiSuNe supports companies in bridging cutting-edge technologies such as Generative AI with practical applications rooted in academic research. Unlike any other new technological paradigm, Generative AI questions the current status quo of work ethics and societal norms.

Therefore, we at AiSuNe guide companies through the complexities of Generative AI adoption and prepare them for the profound changes this technology introduces to workplace dynamics and societal structures. We foster a comprehensive approach to AI integration by addressing both technological and human aspects.

\vspace{1em}
\textbf{Dr. Wolfgang Rohde}\\
Executive Director\\
AiSuNe Foundation\\
\url{https://www.linkedin.com/in/wprohde/}

\end{document}